\documentclass{uai2023} 

\usepackage[american]{babel}

\usepackage{natbib} 
\usepackage{mathtools} 
\usepackage{booktabs} 
\usepackage{tikz} 

\usepackage[utf8]{inputenc} 
\usepackage[T1]{fontenc}    
\usepackage{hyperref}       
\usepackage{url}            
\usepackage{booktabs}       
\usepackage{amsfonts}       
\usepackage{nicefrac}       
\usepackage{microtype}      
\usepackage{xcolor}         

\usepackage{lipsum}

\usepackage{graphicx}
\usepackage{subfigure}
\usepackage{amsmath}
\usepackage{amssymb}
\usepackage{mathtools}
\usepackage{amsthm}

\title{Bag of Policies for Distributional Deep Exploration}

\author[1]{Asen~Nachkov}
\author[1]{Luchen~Li}
\author[1]{Giulia~Luise}
\author[1]{Filippo~Valdettaro}
\author[1,2,3]{Aldo~Faisal}
\affil[1]{%
    Department of Computing\\
    Imperial College London\\
    London, UK
}
\affil[2]{%
    Department of Bioengineering\\
    Imperial College London\\
    London, UK
}
\affil[3]{%
    Lehrstuhlinhaber f\"ur Digital Health \& Data Science\\
    University of Bayreuth\\
    Bayreuth, Bavaria, Germany
  }
  
\begin{document}
\maketitle

\begin{abstract}
Efficient exploration in complex environments remains a major challenge for reinforcement learning (RL). Compared to previous Thompson sampling-inspired mechanisms that enable temporally extended exploration, i.e., deep exploration, we focus on deep exploration in distributional RL. We develop here a general purpose approach, Bag of Policies (BoP), that can be built on top of any return distribution estimator by maintaining a population of its copies.
BoP consists of an ensemble of multiple heads that are updated independently. During training, each episode is controlled by only one of the heads and the collected state-action pairs are used to update all heads off-policy, leading to distinct learning signals for each head which diversify learning and behaviour.
To test whether optimistic ensemble method can improve on distributional RL as did on scalar RL, by e.g. Bootstrapped DQN, we implement the BoP approach with a population of distributional actor-critics using Bayesian Distributional Policy Gradients (BDPG). The population thus approximates a posterior distribution of return distributions along with a posterior distribution of policies. Another benefit of building upon BDPG is that it allows to analyze global posterior uncertainty along with local curiosity bonus simultaneously for exploration. As BDPG is already an optimistic method, this pairing helps to investigate if optimism is accumulatable in distributional RL. Overall BoP results in greater robustness and speed during learning as demonstrated by our experimental results on ALE Atari games.
\end{abstract}

\section{Introduction}



Distributional RL (DiRL)  has rapidly established its place among reinforcement learning (RL) algorithms \cite{dist_persp} as a powerful improvement over non-distributional value-based counterparts \cite{comparison}. In DiRL, the agent does not learn a single summary statistic of the return for each state-action pair, but instead learns the whole return distribution. The agent's behaviour is being evaluated for multiple possible consequences which in turn affect the policy update.
While this does lead to more stable learning and better performance \cite{comparison}, it does not itself change the way actions are selected; as distributional extensions to value-based RL, in C51 \cite{dist_persp}, QR-DQN \cite{qr_dqn} the agent still takes actions according to the mean of the estimated return distributions in each state-action pair. Thus, estimating a return distribution, at its core, provides performance advantages from better representation and evaluation which are unrelated to the action-selection and the exploration behaviour of the agent.

\begin{figure*}
\centering
\includegraphics[width = 0.7\textwidth, height=9cm]{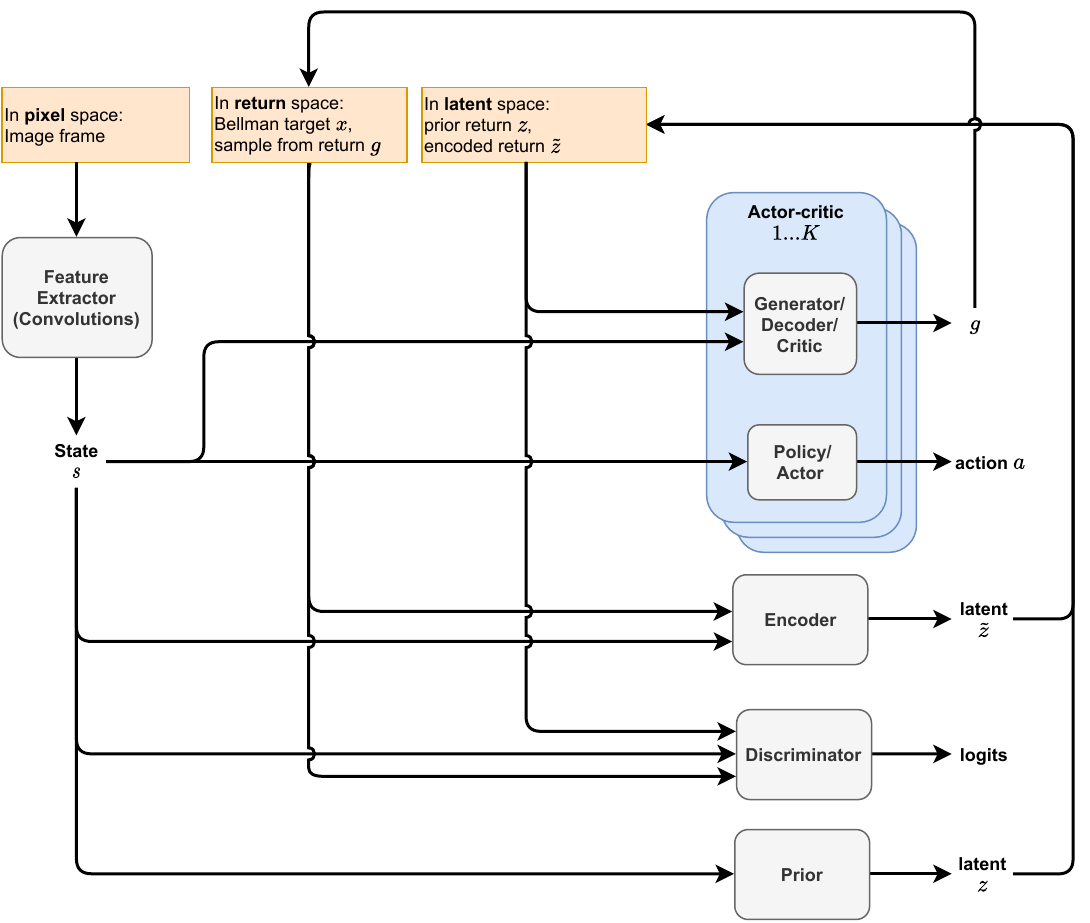}
\caption{Schematic and data flow in BoP.\label{fig:flowbop}}
\end{figure*}

One of the major challenges in DiRL is to better leverage an estimated return distribution for action selection \cite{iqn}. In a discrete action setting, the agent could take the action maximizing the mean plus a time-dependent scaled standard deviation of the return \cite{mavrin}. This approach, combined with computing the upper variance as an embodiment of the ``optimism in the face of uncertainty'', has yielded considerable improvements in exploration. A DiRL solution based on Wasserstein-GAN \cite{freirich19} captures uncertainty in the gradient magnitude of parameters. In the recent Bayesian Distributional Policy Gradients (BDPG) \cite{bdpg}, a curiosity bonus in the form of an information gain is added to the reward function, which motivates the agent to choose actions whose outcome has high epistemic uncertainty in terms of the return distribution estimation. These strategies all bias the data-collecting policy with a curiosity component computed from the return distribution modelling, instead of using only the mean returns as done in conventional RL.

In the meantime, one of the most effective exploration approaches centred around the idea of optimism in the face of uncertainty in scalar RL, is those based on posterior sampling or Thompson sampling \cite{Thompson35}, as we shall discuss in more detail in the Related Work section. These methods typically maintain an ensemble of randomized copies of the same function, relying on model diversity to select actions that are either relatively certainly optimal or uncertain, and therefore efficiently expand the data space with maximal information gathering.

In this work, we want to combine ensemble method with already optimistic DiRL to see if extra optimism from diversified model estimation is beneficial. To this end, we enhance BDPG \cite{bdpg} into multiple distributional actor-critic models to achieve a combination of both Thompson sampling and curiosity bonus induced optimism. Crucially, we are now able to leverage the epistemic uncertainty of the agent in a two-pronged manner: First, through the Thompson sampling enabled by an ensemble of policies and return distributions (heads); Second, by using a local information gain inferred from the observed inaccuracies in each head's return distribution estimate. 

Our contributions in this paper are as follows: we introduce the Bag of Policies framework, analyze its properties and the off-policy training aspects. We also explore variations of our algorithm and test its performance on the Atari Arcade Learning Environment \cite{atari_ale}. Our results suggest that Thompson sampling can indeed improve upon already optimistic DiRL. This finding is promising, as we now know that optimism is accumulatable in DiRL.

\section{Preliminaries}

A reinforcement learning (RL) task is modelled as a Markov decision process (MDP) ($\mathcal{S}, \mathcal{A}, \mathcal{R}, P, \gamma$) \cite{puterman} where $\mathcal{S}$ is the set of possible states of the environment, $\mathcal{A}$ is the set of  actions the agent can take, $\mathcal{R}$ is a reward function , $P$ is the transition probability density for transitioning to a new state given the agent's current state and their selected action, and $\gamma$ is the discount factor. States and actions can be continuous in our framework.
The agent also learns or controls a possibly probabilistic policy $\pi: \mathcal{S} \rightarrow P(\mathcal{A})$ mapping from states to a distribution over actions.

In our implementation the policy $\pi$ is parameterized and updated with policy gradient \cite{sutton, a3c, gae}, maximizing an explicit objective function of the performance of the agent such as $\mathbb{E}_{s \sim d^\pi, a \sim \pi(\cdot | s)}\big[A(s, a) \log \pi(a | s)\big]$, where $d^\pi$ is the stationary marginal state density induced by $\pi$ and $A(s, a)$ the advantage function.

The return for a given state $s\in\mathcal{S}$ is defined as the sum of discounted future rewards that the agent collects starting from the state $s$, $G^\pi(s) := \sum_{t=0}^\infty \gamma^t r_t, \ s_0 = s$.
We consider state-dependent return is this works, while the idea is applicable also to action-dependent return.
The return is a random variable with potential aleatoric uncertainty (e.g. from stochastic state transitions). DiRL methods learn not only a single statistic of the return in a given state, but a representation of the whole distribution. Similar to the Bellman operator defined in conventional RL with respect to the mean of the return, one can use the distributional Bellman operator \cite{dist_persp}. 
\begin{equation}
    \mathcal{T}^\pi G^\pi(s):\stackrel{D}{=}R(s)+\gamma G^\pi(s'),
\end{equation}
where $R(s)$ is the random variable of the reward and $s' \sim P(\cdot|a, s), a\sim \pi(\cdot|s)$, to learn the state return distribution.

We test our idea of improving optimistic DiRL using (also optimistic) deep exploration on the DiRL method BDPG \cite{bdpg}, as it provides an optimistic exploration bonus.
BDPG interprets the distributional Bellman operator as a variational auto-encoding (VAE) process. 
Meanwhile, this VAE provides an estimate of information gain between samples from target and model return distributions, giving rise to the quantification of the epistemic uncertainty in return distribution modelling around the current state under the current policy. This information gain is then used to augment reward signals to encourage exploration. As an already (locally) optimistic DiRL method, BDPG is a good baseline to pair with Thompson sampling for our investigation of buildable optimism in DiRL.

We introduce the superscript $i$ in the notation $\pi^i$ to denote the $i$-th policy in an ensemble of $K$ distributional estimator actor-critic heads.
Accordingly, $A_t^i$ is the advantage of the $i$-th policy at timestep $t$. We adopt the notations used in BDPG to shorthand the model output state-return as $g(s):= G^\pi(s)$, and its Bellman backup target as $x(s):= \mathcal{T}^\pi G^\pi(s)$. They both belong to the return space. BDPG matches $g(s)$ to $x(s)$ in Wasserstein metric (under which the distributional Bellman operator is known to be a contraction \cite{dist_persp}) with a deterministic decoder $p_\theta(x|z, s)$ that maps a latent random variable $z$ to a return sample $g(s)$, a variational encoder $q_\phi(z|x, s)$ that approximates the posterior over $z$ conditional on observed data $x(s)$, and a discriminator $D_\psi(x, z, s)$ to enforce joint-contrastive learning \cite{adversarial_features} now that the decoder is non-differentiable. For algorithm details please refer to the original paper \cite{bdpg}.
Dependency on $s$ is omitted when no confusion is to be incurred. We adapt per-head shorthand from BDPG notations for the Bellman targets as $x_t^i$, the samples from the generators $g_t^i$, and latent variables for the generative process $z_t^i$.

\section{Related Work}
Closest to our work is the multi-worker or multi-head approaches which leverage Thompson sampling \cite{Thompson35} to diversify behaviour. Notice that unlike parallel approaches such as A3C \cite{a3c} and variants \cite{liang2018gpu,zhang2019asynchronous} in which the multiple workers are updated with the same gradient, Thompson sampling allows the workers to learn from bootstrapped signals \cite{bootstrap} and to focus on where the current understanding is insufficient rather than merely cover more state space more quickly together than a single worker.

One early example of Thompson sampling in deep RL is the Bootstrapped DQN \cite{bootstrapped_dqn} which utilizes multiple $Q$-value ``heads''. Each head is initialized randomly and differently and then trained on their respective learning signal (with slightly different data or even identical data but distinct gradient). Thanks to this diversity, 
the heads' evaluations for a state-action pair tend to converge to the true value (which then allows the optimal action to be chosen) as the pair is being sufficiently visited. Thus said, an action is selected either because it is optimal or because its evaluation is uncertain, thus optimistic. This uncertainty is referred to as the epistemic uncertainty that would diminish as training data increases. Bootstrapped DQN and variants \cite{osband19, chen17ucb, odonoghue18} can be thought of as representing a posterior distribution of the $Q$-function and thus gives a measure of epistemic uncertainty in the $Q$-value estimation. Similarly, \cite{tang} approximates Thompson sampling by drawing network parameters for an RL model. To further encourage optimism in the face of uncertainty, gradient diversification \cite{diversifiedQ21} can effectively boost performance of an ensemble method.

Ensemble method is important as it also allows for deep exploration by engaging the same policy for the duration of the whole episode. However, all ensemble-based exploration schemes for now are intended for conventional scalar RL, and the purpose of this work is to investigate whether deep exploration could improve performance in distributional RL, further still, whether the global optimism incurred by the ensemble effect can further enhance exploration efficiency when the agent is already optimistically biased on per-state basis. Combining ensemble technique with DiRL is not a trivial investigation and the benefit of diversity does not automatically propagate, as now we would be learning a posterior distribution over \emph{distributions}, whilst admittedly the diversity over distributions is harder to maintain than that over numbers.

We base our approach on BDPG \cite{bdpg} which is a DiRL approach made up of a single distributional actor-critic and provides a local curiosity bonus favouring states whose return distribution estimations suffer from high epistemic uncertainty. BoP in contrast maintains a posterior distribution over return distributions and over policies, keeping track of both aleatoric and epistemic uncertainties about return distribution estimation simultaneously.

\section{Bag of Policies (BoP)}
In the following we lay out the structure, theory and variants of BoP. A schematic can be found in Fig.~\ref{fig:flowbop} and for the pseudocode please refer to Algorithm$~1$.

\begin{algorithm}[!htb]
    \caption{Bag of Policies}
    \label{bop_pseudocode}
\begin{algorithmic}[1]
    \STATE Initialize prior $p_\theta(z|s)$, encoder $q_\phi(z|x, s)$, discriminator $D_\psi(x, z, s)$, actor-critic population \\$\{G_\theta^i(z, s),  \pi^i\}_{i=1}^K$
    \WHILE{not converged}
    \STATE // Roll-out stage
    \STATE training batch $\mathcal{D} \gets \emptyset$
    \STATE Sample $k \sim \mathrm{Uniform}(1, ..., K)$ // actor-critic to act
    \FOR{$t=0$ to $T-1$} 
    \STATE execute $a_t^k \sim \pi^k(\cdot|s_t)$, get $r_t, s_{t+1}$
    \STATE sample return $z_t \sim p_\theta(\cdot|s_t),  g_t^i \gets G_\theta^i(z_t, s_t), \  \forall i$
    \STATE $\mathcal{D} \gets \mathcal{D} \cup \big(s_t, a_t, r_t, g_t^i, \log \pi^i (a_t | s_t)\big), \forall i$
    \ENDFOR
    \STATE sample last return \newline $z_T \sim p_\theta(\cdot|s_T),  g_T^i \gets G_\theta^i(z_T, s_T),  \ \forall i$
    \STATE // Data-driven estimation stage
    \FOR{$t$ $\in$ $\mathcal{D}, ~\forall i$}
    \STATE estimate off-policy advantage $A_t^i$
    \STATE compute Bellman target $x_t^i \gets A_t^i + g_t^i$
    \STATE compute mixed curiosity reward \newline $\hat{r}_t^i \propto \text{KL}\big(q_\phi(\cdot | x_t^k, s_t) \ || \ q_\phi(\cdot | g_t^i, s_t)\big)$
    \STATE augment $A_t^i$ by replacing $r_t$ with $r_t + \hat{r}_t^i$
    \ENDFOR
    \STATE // Update stage
    \STATE // Train with minibatch $B \subset \mathcal{D}$
    \FOR{$t$ $\in$ $B$}
    \STATE encode $\tilde{z}_t^i \sim q_\phi(\cdot | x_t^i, s_t), ~\forall i$
    \STATE sample adversaries \newline $z_t^i \sim p_\theta(\cdot|s_t), \ \tilde{x}_t^i \gets G_{\bar{\theta}}^i (z_t^i, s_t), \ \forall i$
    \STATE take averages \newline $\bar{z}_t = \frac{1}{K} \sum_{i=1}^K \tilde{z}_t^i,  \bar{x}_t = \frac{1}{K} \sum_{i=1}^K \tilde{x}_t^i$
    \ENDFOR
    \STATE update $D_\psi$ by ascending \newline 
            $\frac{1}{|B|}\sum_{t \in B} \big[\log D_\psi(\bar{x}_t, z_t, s_t) + \newline ~~~~~~~~~~~~~~~~~~~~\log \big(1-D_\psi(x_t, \bar{z}_t, s_t)\big)\big]$
    \STATE update encoder, prior by ascending \newline 
    $\frac{1}{|B|}\sum_{t \in B} \big[\log \big(1 - D_\psi(\bar{x}_t, z_t, s_t)\big) + \newline ~~~~~~~~~~~~~~~~~~~~ \log D_\psi(x_t, \bar{z}_t, s_t )\big]$
    \STATE update $G_\theta^i, \ \forall i$ by descending \newline $\frac{1}{|B|}\sum_{t \in B} ||x_t^i - G_\theta^i(\tilde{z}_t^i, s_t)||_2^2$
    \STATE update $\pi^i, \ \forall i$ by ascending \newline $\frac{1}{|B|} \sum_{t \in B}  \log \pi^i(a_t | s_t)A_t^i$
    \ENDWHILE
    \end{algorithmic}
\end{algorithm}

\subsection{Architecture}

The Bag of Policies (BoP) framework can be applied to any DiRL estimator. In this work we have chosen BDPG \cite{bdpg} as the implementation framework because it exploits local epistemic uncertainty. We use the architecture as that of  BDPG except that we have multiple distributional actor-critic heads.
Specifically, we refer to each distributional actor-critic pair (policy + return distribution) as a head. Each episode of trajectory is rolled out  by only one of the heads sampled uniformly randomly, and the collected data is trained on by all of the heads for data efficiency. The head that generated the trajectory learns on-policy, whilst all other heads learn from the same batch of data off-policy. Throughout training, the heads are updated by different gradients because they were initialized differently and learn from their own Bellman target which entails random sampling from their own return distributions at subsequent steps. This ensures a diversity in policies, giving rise to the effect of Thompson sampling.

\begin{figure*}
\centering
\includegraphics[width=\textwidth]{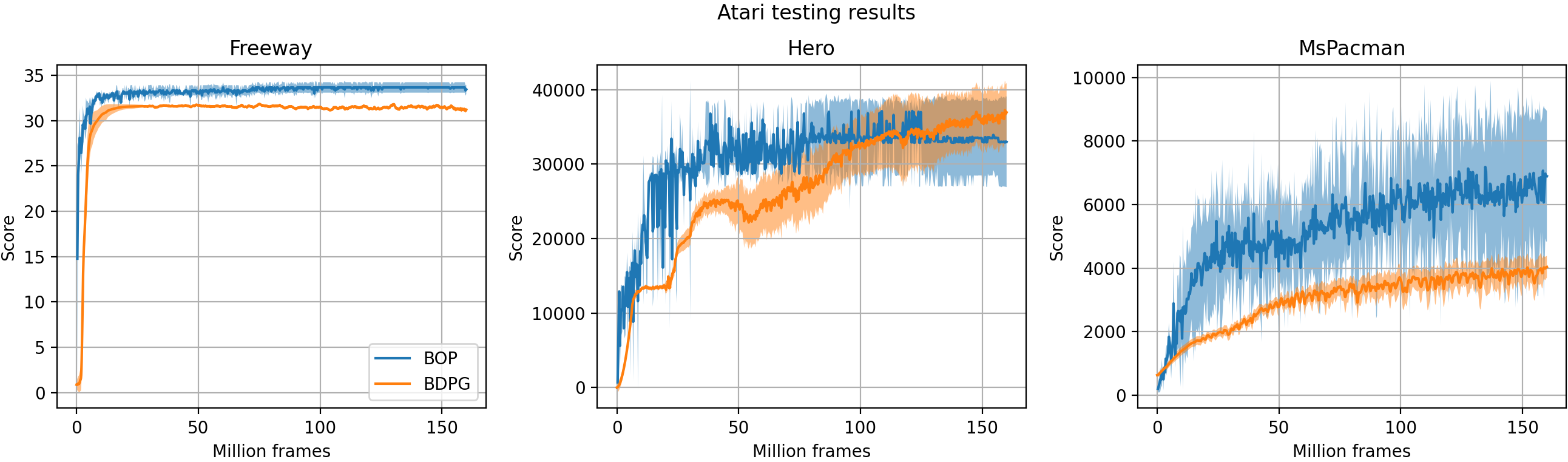}
\caption{Comparison of BDPG and BoP on selected Atari environments.}
\label{fig:BDPGBOP}
\end{figure*}

The training of BoP entails an outer loop of three stages: a roll-out stage, a data-driven estimation stage, and an update stage. The majority of the algorithm is the same as that of BDPG, we only highlight the different or extra aspects incurred by using multiple distributional actor-critic heads, as well as the heed we paid to maintain diversity. During the roll-out stage, at the start of each episode, the agent selects a head uniformly at random, which we call the active head $k$, and executes actions according to the policy of that head for the duration of the whole episode. Thus, the sequence of collected transitions $(s_t, a_t, r_t)$ is determined entirely by the policy of the active head as a behaviour policy. 
Meanwhile, the other heads still compute the log-probability of the selected action $\log \pi^i (a_t | s_t)$, and form an estimation of the return distribution for each state in the episode by generating a sample $g_t^i$ from each distributional critic.
These local evaluations are used to provide them with a unique sample estimate of the learning objective when updating their parameters.

During the sample estimate stage, the agent computes the advantages $A_t^i$ and the Bellman targets $x_t^i$, both being unique for each respective head. These local estimates also engender a local curiosity bonus $\hat{r}_t^i$ that is proportional to the entropy reduction in the latent space conditioned on the return space due to the condition being a more accurate estimate (the on-policy Bellman target $x_t^k$) as opposed to a model prediction (a sample from the model $g_t^i$): 
\begin{align}\label{eq:curiosity_rew}
    \hat{r}_t^i &\propto\mathrm{Information\_gain} (s_t, i) \nonumber\\
    &:= \mathrm{KL}\big(q_\phi(\cdot | x_t^k, s_t) \ || \ q_\phi(\cdot | g_t^i, s_t)\big),
\end{align}
in which the state $s_t$ is generated by the active head $k$.
This entropy reduction, or equivalently, information gain, quantifies the epistemic uncertainty about how well each head is modelling the return distribution. 
The posterior $x_t^k$ is shared across heads as it is the only Bellman backup target among all $x_t^i$ that is estimated on-policy and thus suffers the least degree of bias / variance. The curiosity bonus is then added to the external reward for computing the augmented advantage function that would be used to update the local policies. 

During the update stage, all policies $\pi^i$ are updated on the data generated by the active policy $\pi^k$.
The active head is updated on-policy as in BDPG. For the other heads we use an off-policy advantage function estimated with V-trace \cite{Espeholt2018IMPALA} at each timestep.
Subsequently, each head's critic  is updated by minimizing a Wasserstein distance between its current return distribution model and the distribution of a Bellman target estimated from its own predictions at subsequent timesteps. Like BDPG, this is achieved by minimizing the squared distance between  Bellman target $x_t^i$ and sample $g_t^i$ while in the meantime enforcing joint-contrastive learning adversarially.

We used a single common set of hyperparameters for implementing the BoP on Atari as shown in Table ~\ref{atari_hyperparameters}. The choices for the number of heads were laid out in Table ~\ref{atari_results_table}. The policy heads are updated by Proximal Policy Optimization (PPO) \cite{ppo}.

\begin{table*}[t]
\small
\centering
  \begin{tabular}{l l l l}
    \toprule
    \textbf{Hyperparameter} & \textbf{Value} & \textbf{Hyperparameter} & \textbf{Value}\\
     \midrule
     Num. parallel envs  & 16 & Evaluation frequency & Every $2 e4$ timesteps\\
     Num. stacked frames  & 4 & Num. heads &  Variable\\
     GAE $\lambda$  & 0.95 & Discount $\gamma$  & 0.99\\
     Num. episodes per roll-out  & 4 & Minibatch size  & 256\\
     Roll-out timesteps  & 128 & Value loss weight  & 0.5\\
     Entropy coefficient  & 0.01 &  Learning rate  & Linearly $2.5 e-4\to 0$\\
     PPO clip range  & Linearly $0.1\to 0$ &  Curiosity & Mixed\\
     Actor-critics updated & All \& on all the data & $x_t^i$ computed by & Critic $i$\\
     Actions during roll-out & Sampled & Actions during testing & Greedy\\
     \bottomrule
  \end{tabular}
    \caption{Hyperparameters for implementing BoP model on Atari games.}
    \label{atari_hyperparameters}
\end{table*}

\subsection{Design considerations \& implications}

In this subsection we discuss the possible variants of BoP enabled by its ensemble characteristic, and which of them are conducive to diversity. To understand these and the design decisions leading to the final settings described above, we first explain the need for diversification and the off-policy side of our framework.

Note that the following considerations for diversification are not entailed in ensemble methods applied to scalar RL, where behaviour and model diversity is much easier to uphold, i.e., a surprising backup target is considered unseen in a 
scalar model but merely less likely in a distribution model. Thus said, the same level of diversity would provide lesser learning signal to the latter. This means if the posterior over return distributions is not sufficiently diverse, which would happen if ensemble techniques and DiRL were naively combined, the return distribution models will only tweak the likelihood for the current backup target rather than fundamentally change its return prediction; and the ensemble members remain similar after light update, forming a vicious cycle.

With multiple heads that are being maintained in BoP, each head $i$ can form a return estimate $g_t^i$ at the current timestep and produce a Bellman target $x_t^i$ by sampling from itself for each of the future timesteps. Thus, there are choices about sharing these targets or not across heads in two domains: 1. to update the critic and 2. to compute the local curiosity bonus.

First we considered sharing the Bellman target $x_t^k$ across critics,  i.e., using the target of the current active head $k$ to update all critics.
We found that in practice sharing Bellman targets causes the diversity between policies to collapse, as measured by various metrics such as the sum of all absolute differences between the assigned probabilities of any two policies and the cosine similarity between two policies. This resulted in poorer exploration outcomes than with individual Bellman targets. 
Therefore, in the final BoP framework, the on-policy Bellman target is not shared and each critic head is trained against its own target, which provides (more) unique bootstrapped signals to each head. Formally, we use the following variation of the distributional Bellman equation:
\begin{align}
    &\mathcal{T}^{\pi^k} G^{\pi^i} (s_t)\stackrel{D}{=}r_t+\gamma G^{\pi^i} (s_{t+1})\\ \nonumber
    &s_{t+1} \sim P(s_{t+1} | s_t, a_t), ~a_t \sim \pi^k(s_t).
\end{align}
Since the states, actions and rewards are collected by the active head $k$, only this head is trained on-policy, while all other heads are trained off-policy. Thus, the final version of BoP is an algorithm of a mix of on-policy and off-policy learning, in contrast to other algorithms \cite{dist_persp, a3c, bdpg, bootstrapped_dqn, qr_dqn}, which are either one or the other.

Each distributional actor-critic member in BoP has a  curiosity bonus $\hat{r}_t^i$ applied to its personal policy. We have discussed the reason for sharing the posterior return estimate $x^k_t$ in Eq. \eqref{eq:curiosity_rew} which is for its accuracy, we here consider the choices for the prior $g_t$. This prior can either be computed from that head itself $\mathrm{KL}\big(q(\cdot | x_t^k, s_t) \ || \ q(\cdot | g_t^i, s_t)\big)$ as in Eq. \eqref{eq:curiosity_rew} or shared $\mathrm{KL}\big(q(\cdot | x_t^k, s_t) \ || \ q(\cdot | g_t^k, s_t)\big)$, where $q(z|x, s)$ is the variational encoder for the return distribution. We use local return samples $g_t^i$ as prior to compute the information gain. This is because a sample surprising to one return distribution model might not surprise another, and each model should keep track of its accuracy to correctly bias local action selection, so that the model agreement on policies informs also implicitly the agreement on the level of surprise.

Next, we consider the schedule for updating heads: we can update all of them on every batch of freshly collected data, or just some of them. It is indeed feasible to update only one head - e.g. the active one or the most uncertain one, as measured by the curiosity bonuses. By updating only the most uncertain head at each iteration, the agent maximizes the reduction in epistemic uncertainty while still keeping the heads sufficiently diverse (by updating only one head in each update iteration). However, we found in practice that this diminishes sample efficiency due to limiting the number of heads being trained with a given amount of data. Similar empirical findings were made when training with other setups that selectively update members like updating a randomly sampled head according to its epistemic uncertainty, or training each head on 50\% of their most uncertain samples. For that reason in BoP we update all heads on all the data, which provides the fastest learning at no cost of sacrificing sufficient diversity of the heads and therefore maximizes their ensemble benefits on balance.

During testing, on the other hand, the question arises as to how action selection should be performed now that we have multiple policies. BoP selects an action by averaging the action distributions of all policy heads and then chooses the greedy action from that distribution \cite{ensemble}. Since all policies in our implementation are Gaussian, this is easy to realize.
In practice we observed more stable performance and higher scores when using this ``average-then-argmax'' approach compared to other options, such as each head picking their greedy action and then performing a majority vote  selection of the most popular / similar action across heads (``argmax-then-vote''), or even selecting actions only from one randomly selected head (``sample-then-argmax''). 
The benefits of the ``average-then-argmax'' tactic over the others result from relying on model agreement which is what has been truly learned and understood rather than uncertainty-quenching exploration.

Finally, we considered how to choose the number of ensemble members. We found that a small number of $3$ to $5$ strikes a good balance between effective exploration and training speed. For calibration of this statement, we found that each additional head adds about $15\%$ more FLOPS to the total amount of computation. This increase in performance from having many heads saturates as policies start to converge and thus become more and more correlated, which justifies using a small number of heads to increase performance and exploration efficiency without increasing training time too much.

\begin{figure}[t]
\centering
\includegraphics[width=\columnwidth]{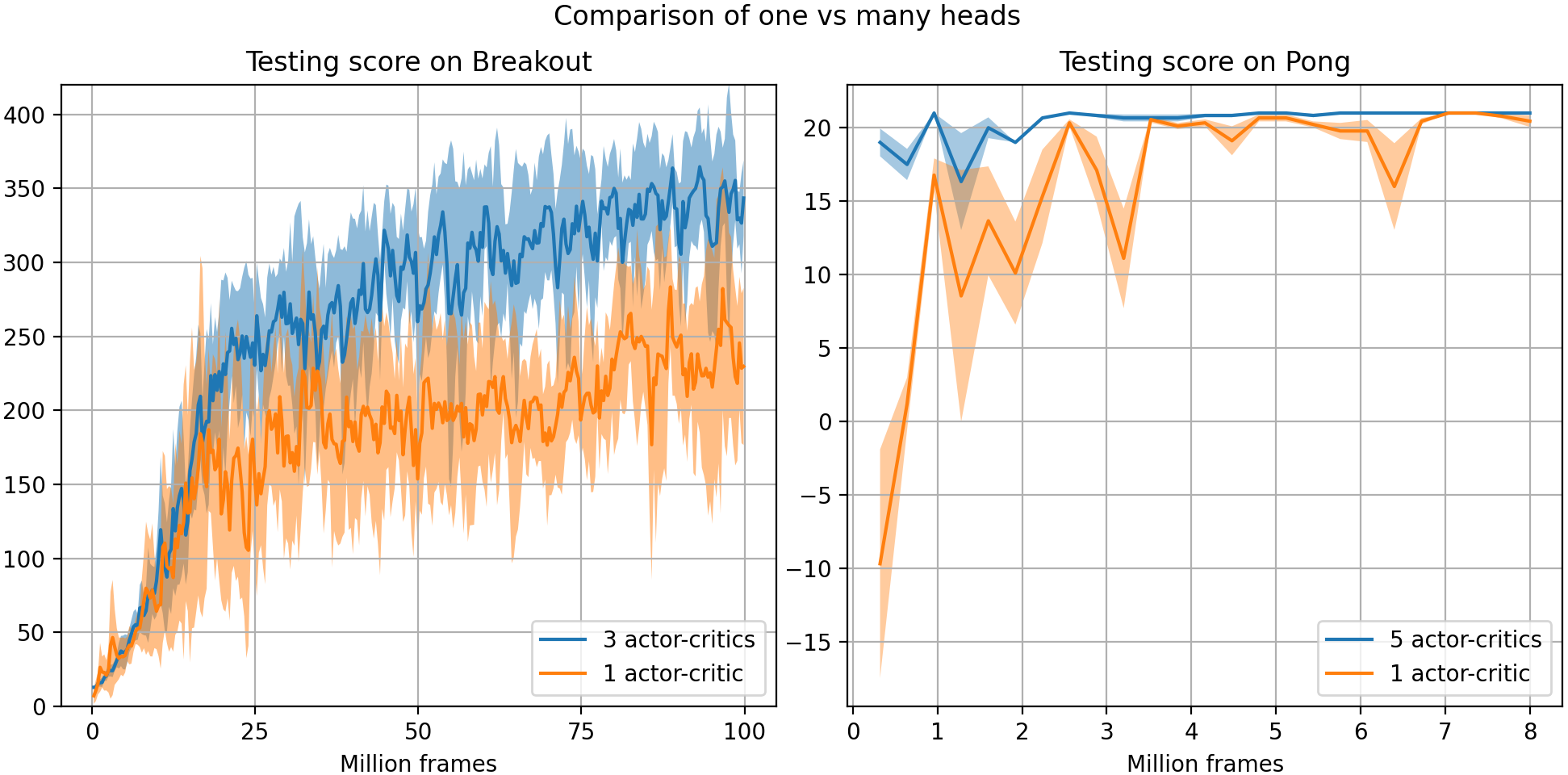}
\caption{Comparison of BoP with one $(K=1)$ vs many heads $(K=3)$ or $(K=5)$ on Breakout and Pong. The shading shows one standard error from the mean. Exponential smoothing was applied for the scores on Breakout. The numbers of heads in multi-head cases shown are the minimal numbers that enable significant improvement over one-head baseline. \label{fig:heads}}
\end{figure}

\section{Experimental results}

We tested the performance of BoP on a diverse set of Atari environments, and compared to three baselines with which BoP shares similarity in different aspects: to Bootstrapped DQN \cite{bootstrapped_dqn} as an ensemble-based exploration in scalar RL, to BDPG \cite{bdpg} as a locally optimistic DiRL exploration approach and the algorithm we built upon, and to A3C \cite{a3c} which facilitates covering extensive data space mainly by deploying multiple workers in parallel.
The set of Atari environments were chosen to represent an exemplar mix of hard exploration environments, mazes, and shooter games. Pixel-based observations were pre-processed by standard Atari wrappers as laid out in \cite{qr_dqn}, including cropping the image frame, using only grayscaled frames, frame stacking, and taking the maximum values of any two consecutive frames.

First of all, we compare performances between BDPG and BoP on $\mathrm{Freeway}$, $\mathrm{Hero}$ and $\mathrm{MsPacman}$, with learning curves shown in Fig.~\ref{fig:BDPGBOP}. 
Compared to BDPG, BoP improves significantly on sample efficiency, albeit not necessarily on asymptotic performance.
We note that the only difference between BDPG and BoP is that BoP is a bootstrapped ensemble of BDPG, so the improvement of learning speed is in principle attributed to the ensemble effect. However, at this stage, it cannot be decided it is the deep exploration enabled by Thompson sampling or the sheer multitude of workers that has resulted the advantage of ensemble technique.

\begin{table}[t]
\small
\centering
  \begin{tabular}{|l||r|r||r|r||r|r|}
    \hline
    \textbf{Atari env} & \textbf{\#} & \textbf{BoP} & \textbf{\#}  & \textbf{A3C} 
    & \textbf{\#}  & \textbf{Boot} \\
    & & & & &
    & \\
     \hline
     Freeway    & 3 & \textbf{34} & 16 & 0.1 & 10 & \textbf{34} \\
     \hline
     Breakout  &3 & 686 & 16 & 682 & 10 & \textbf{855} \\
     \hline
     Hero & 3 & \textbf{37,728} & 16 & 32,464 & 10 & 21,021\\
     \hline
     MsPacman  & 3 & \textbf{9,057} & 16 & 654 & 10 & 2,983 \\
     \hline
     Qbert  & 3 & \textbf{20,583} & 16 & 15,149 & 10 & 15,092 \\
     \hline
     Alien  & 5 & \textbf{5,470} & 16 &  518 & 10 & 2,436  \\
     \hline
     Asterix  & 5 & \textbf{42,500} & 16 &  22,141 & 10 & 19,713 \\
     \hline
     Frostbite  & 3 & \textbf{5,940} & 16 & 191 & 10 & 2,181 \\
     \hline
     Amidar  & 5 & 1,123 & 16 & 264 & 10 & \textbf{1,272}  \\
     \hline
     BeamRider & 3 & 6,499 & 16 & 22,708 & 10 & \textbf{23,429}  \\
     \hline
     StarGunner  & 3 & 56,200 & 16 & \textbf{138,218} & 10 & 55,725 \\
     \hline
     Seaquest  & 3 & 2,027 & 16 &  2,355 & 10 & \textbf{9,083} \\
     \hline
  \end{tabular}
    \caption{The testing results achieved across 3 runs. The number of heads (``\#'') for BoP were set by us, the ones for A3C and Bootstrapped DQN (``Boot'')   were set in their original works \cite{a3c,bootstrapped_dqn}.}
    \label{atari_results_table}
\end{table}

To this end, we move on to investigate the impact of the number of ensemble members, i.e. heads, on a multi-worker agent's performance. In Table ~\ref{atari_results_table}, we compare BoP against multi-worker / multi-head methods Bootrapped DQN and A3C. Although BoP does not beat baselines on all selected environments, we can see that it indeed outperforms the others most frequently. Crucially, the  number of heads for A3C ($K=16$) and Bootstrapped DQN ($K=10$), as proposed by their original publications, are substantially higher than those needed for BoP ($K=3$ or in two cases $K=5$). Moreover, for all games where deep exploration is essential (such as $\mathrm{MsPacman}$ and $\mathrm{Frostbite}$), BoP substantially outperforms baselines by $50\% - 200\%$. These promising results suggest that increasing the multitude of workers alone is not key to better performance. Combining with the observations made in comparing to BDPG, one can establish that the strengths of BoP are attributed to deep exploration. In the meantime, we also observed that BoP is not as competitive in games with moving targets but not exploration-demanding, such as $\mathrm{BeamRider}$ and $\mathrm{Seaquest}$. We surmise that this is because BoP became overly explorative seeing the moving subjects whereas the motion could not be accounted for by being bold unlike in the case of $\mathrm{Freeway}$.

\begin{figure}[tb]
\centering
\includegraphics[width=\columnwidth]{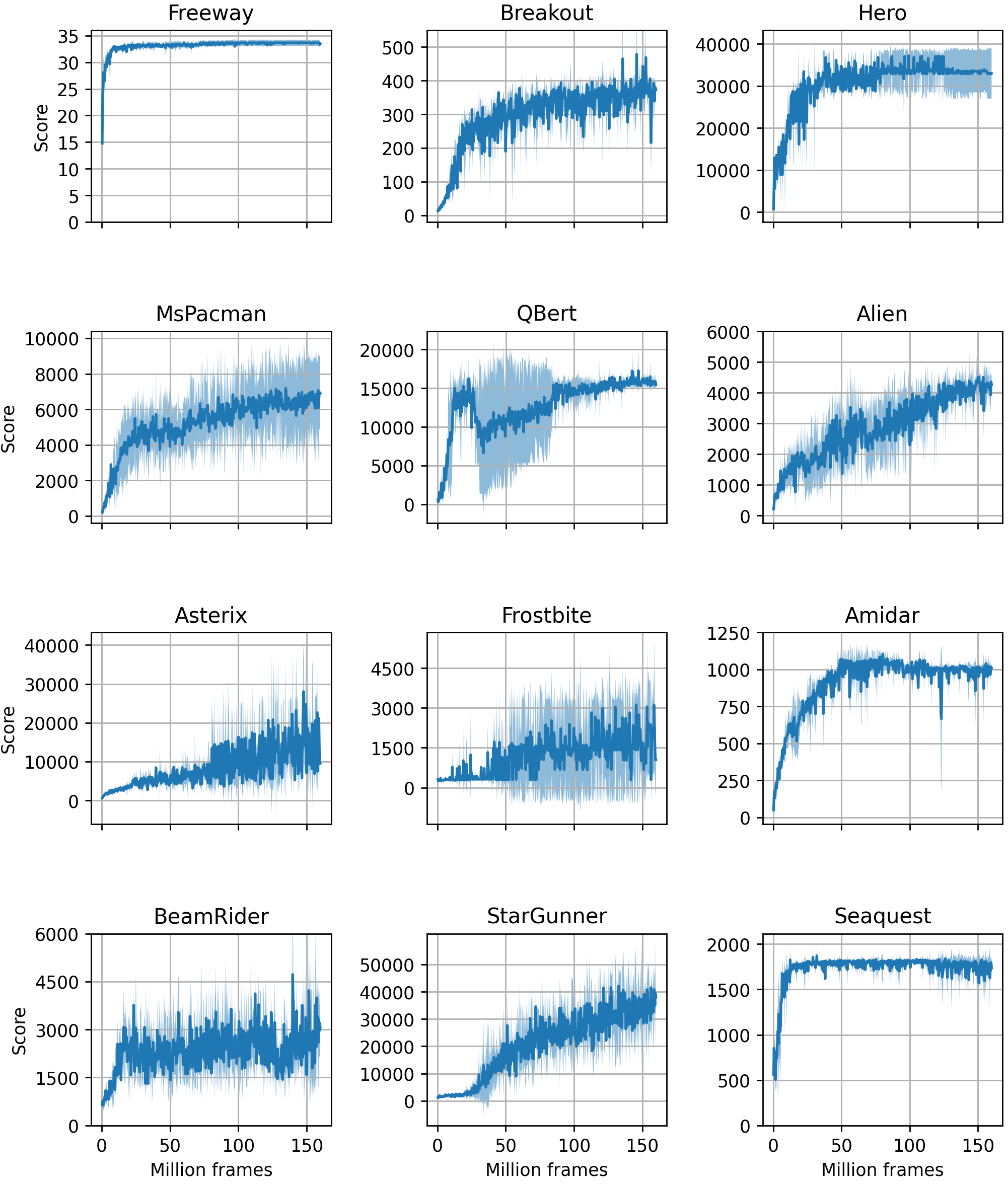}
\caption{Testing scores for BoP on Atari. We test the agent periodically while training. Shaded areas indicate variation in performance across multiple runs.\label{fig:learningcurves}}
\end{figure}

Another experiment that we did to investigate the ensemble effect is to compare a one-head BoP against a few-head BoP (Fig.~\ref{fig:heads}). We noted a substantial improvement when the number of heads increased from one to even a very small value (e.g. $3$), while being a qualitative transformation from a flat algorithm into an ensemble one. This finding reinforces the notion that the benefit of BoP is not due to merely more workers, but a fundamentally distinct way that an ensemble method works in -- by posterior sampling and therefore favouring uncertain areas for exploration.

This performance improvement outweighs the increase in compute time for a few-head BoP. This is probably due to the better exploration capability of each head operating with an independently updated policy learning from its own estimation errors so that an action is selected either because it is agreed to be relatively certainly optimal or is uncertain. However, when the number of ensemble members becomes too large, BoP will have lost its advantage. As mentioned earlier, we observed that on Atari each additional head adds about $15\%$ more FLOPS for the full duration of the training compared to having only one head. This linear relationship holds true also when having $10$ or $15$ heads. The marginal cost in terms of FLOPS is constant. The marginal benefit from adding another head decreases with the number of heads quickly, resulting in our lower use of heads than other multi-worker algorithms. In conclusion, BoP is most advantageous when using a handful of members.

The learning curves of all Atari environments that BoP was tested on can be found in Fig.~\ref{fig:learningcurves}. For instance, $\mathrm{Frostbite}$ is a game that needs very specific sequences of actions, similar to maze games like $\mathrm{MsPacman}$ or $\mathrm{QBert}$ that require deep exploration, as decisions which one takes early on have long term impacts.

\section{Discussion \& future work}

The Bag of Policies (BoP) algorithm presented here is a multi-head distributional estimator applicable to a large class of algorithms and thus extends deep exploration to distributional RL settings. Our most salient finding is that the benefits of the optimism in the face of uncertainty \emph{can accumulate} for DiRL exploration, as we were able to substantially improve upon a DiRL approach that is optimistic on per-state basis simply by making an ensemble of it and performing Thompson sampling. This suggests that at least in DiRL, Thompson sampling and curiosity bonus, although both facilitate optimism, can work in conjunction to enhance performance further.

In addtion, as our experimental results showed, the advantage of BoP is not attributed to the mere multitude of workers as BoP can surpass other multi-worker / multi-head algorithms with much fewer heads, and to boost performance considerably with a few-head BoP from a one-head counterpart. These experiments substantiate that the ensemble technique which empowered BoP functions in a fundamentally different fashion than naively summing up the efforts of parallel workers -- by Thompson sampling, which enables optimistic and deep exploration in the same time.

On average, BoP achieves better learning speed and asymptotic performance than baselines.
We have observed the biggest improvement from baselines in maze-like games like $\mathrm{MsPacman}$ or $\mathrm{QBert}$ where the agent has to choose a path in a labyrinth while collecting various items scattered throughout it and avoiding the enemies many moves down the line. These kinds of environments require deep exploration and the agent can experience vastly different outcomes depending on which path it takes. Hence, these environments provide a good example where the BoP exploration capabilities can improve the agent's performance.

In a nutshell, BoP has demonstrated that not only deep exploration is viable in DiRL, offering extensions to any DiRL settings (e.g. \cite{dist_persp, qr_dqn, iqn, freirich19, doan18, martin20, barthmaron18, Singh20, kuznetsov20, Choi19}), but also that deep exploration can further improve learning from an already optimistic exploration strategy.

\subsubsection*{Acknowledgments}

We are grateful for our funding support. At the time of this work, GL and AF are sponsored by UKRI Turing AI Fellowship (EP/V025449/1), LL and FV by the PhD sponsorship of  the Department of Computing, Imperial College London.


\end{document}